\documentclass[11pt,letterpaper]{article}
\usepackage{emnlp2016}
\usepackage{times}
\usepackage{amssymb}
\usepackage{latexsym}
\usepackage{subfigure} 
\usepackage{multirow}
\usepackage{url} 

\usepackage{breakurl}
\usepackage[utf8]{inputenc}
\usepackage{graphicx}
\usepackage{array}
\usepackage{amsmath}
\usepackage[font=singlespacing,small]{caption}
\usepackage{fancyhdr}
\usepackage{amssymb}
\usepackage{nccmath}
\usepackage{color}
 \usepackage[normalem]{ulem}
 \usepackage{lipsum}
 \usepackage{graphics}
 \usepackage{tikz-dependency}
 \usepackage{enumitem}
 
\setitemize{noitemsep,topsep=10pt,parsep=0pt,partopsep=0pt}
\setenumerate{noitemsep,topsep=10pt,parsep=0pt,partopsep=0pt}

 \newcommand{\ignore}[1]{}

 \newcommand{\adhicomment}[1]{\ignore{\textcolor{green}{{\textbf{[#1 --\textsc{adhi}]}}}}}
 \newcommand{\lingpengcomment}[1]{\ignore{\textcolor{green}{{\textbf{[#1 --\textsc{lpk}]}}}}}
 \newcommand{\cjd}[1]{\textcolor{cyan}{\ignore{{\textbf{[#1 --\textsc{cjd}]}}}}}
 \newcommand{\nascomment}[1]{\ignore{\textcolor{blue}{{\textbf{[#1 --\textsc{nas}]}}}}}
 \newcommand{\miguelcomment}[1]{\ignore{\textcolor{red}{{\textbf{[#1 --\textsc{miguel}]}}}}}

\makeatletter
\newcommand*\bigcdot{\mathpalette\bigcdot@{.5}}
\newcommand*\bigcdot@[2]{\mathbin{\vcenter{\hbox{\scalebox{#2}{$\m@th#1\bullet$}}}}}
\makeatother

\emnlpfinalcopy

\def\emnlppaperid{820}

\title{Distilling an Ensemble of Greedy Dependency Parsers into One MST Parser
    }

\author{Adhiguna Kuncoro$^{\spadesuit}$ ~ Miguel Ballesteros$^{\diamondsuit}$  ~ Lingpeng Kong$^{\spadesuit}$  ~ Chris Dyer$^{\spadesuit\clubsuit}$ ~ Noah A. Smith$^{\heartsuit}$ \\
$^{\spadesuit}$School of Computer Science, Carnegie Mellon University, Pittsburgh, PA, USA \\
$^{\diamondsuit}$NLP Group, Pompeu Fabra University, Barcelona, Spain \\
$^{\clubsuit}$Google DeepMind, London, UK\\
$^{\heartsuit}$Computer Science \& Engineering, University of Washington, Seattle, WA, USA\\
{\small \tt \{akuncoro,cdyer,lingpenk\}@cs.cmu.edu}\\ {\small \tt miguel.ballesteros@upf.edu, nasmith@cs.washington.edu}
}

\date{}

\begin{document}

\maketitle

\begin{abstract}
We introduce two first-order graph-based dependency parsers achieving a new state of the art. The first is a consensus parser built from an ensemble of independently trained greedy LSTM transition-based parsers with different random initializations. We cast this approach as minimum Bayes risk decoding (under the Hamming cost) and argue that weaker consensus within the ensemble is a useful signal of difficulty or ambiguity. The second parser is a ``distillation'' of the ensemble into a single model. We train the distillation parser using a structured hinge loss objective with a novel cost that incorporates ensemble uncertainty estimates for each possible attachment, thereby avoiding the intractable cross-entropy computations required by applying standard distillation objectives to problems with structured outputs.\ignore{\cjd{Rather than giving the accuracies here, I think I'd prefer to say something about the distillation cost objective: maybe, it is ``the Hamming cost discounted by the posterior (or `ensemble'?) attachment uncertainty estimated from the ensemble of predictions.'' No one outside of parsing will remember or care about our numbers, people might care about this idea. I think including the parsing numbers makes us seem parochial, and this is a more interesting paper.}}
The first-order distillation parser matches or surpasses the state of the art on English, Chinese, and German. 
\end{abstract}
\section{Introduction}

Neural network dependency parsers achieve  state of the art performance
\cite{stack_lstm,structured_training,globally_normalized}, but training them
involves gradient descent on non-convex objectives, which is unstable with respect to initial parameter values.  For some tasks, an \textbf{ensemble} of neural networks from different random initializations has been found to improve performance over individual models \cite[\emph{inter alia}]{seq_to_seq,grammar_foreign}.  In \S\ref{sec:mbr}, we apply this idea to build a first-order graph-based (FOG) ensemble parser \cite{ensemble_reparsing} that seeks consensus among 20 randomly-initialized stack LSTM parsers \cite{stack_lstm}, achieving nearly the best-reported performance on the standard Penn Treebank Stanford dependencies task (94.51 UAS, 92.70 LAS).

We give a probabilistic interpretation to the ensemble parser (with a minor modification), viewing it as an instance of \textbf{minimum Bayes risk} inference.  We propose that disagreements among the ensemble's members may be taken as a signal that an attachment decision is difficult or ambiguous.

Ensemble parsing is not a practical solution, however, since an ensemble of $N$ parsers requires $N$ times as much computation, plus the runtime of finding consensus.  We address this issue in \S\ref{sec:distillation} by \textbf{distilling} the ensemble into a \textbf{single} FOG parser with discriminative training by defining a new \emph{cost function},
inspired by the notion of ``soft targets'' \cite{dark_knowledge}.  The essential idea is to derive the cost of each possible attachment from the ensemble's division of votes, and use this cost in discriminative learning.
The application of distilliation to structured prediction is, to our knowledge, new, as is the idea of empirically estimating cost functions.

The distilled model performs almost as well as the ensemble consensus and much better than (i)  a strong LSTM FOG parser trained using the conventional Hamming cost function, (ii) recently published strong LSTM FOG parsers \cite{kiperwasser,graph_based_segment}, and (iii) many higher-order graph-based parsers \cite{third_order,turbo_parser,le_zuidema}.
\adhicomment{Added a citation to the recent graph-based LSTM ACL 2016 paper}
\nascomment{collapsed to go with Kiperwasser}  It represents a new state of the art for graph-based dependency parsing for English, Chinese, and German. 
The code to reproduce our results is publicly available.\footnote{\url{https://github.com/adhigunasurya/distillation_parser.git}} 

\section{Notation and Definitions}

Let $\boldsymbol{x} = \langle x_1, \ldots, x_n\rangle$ denote an $n$-length sentence.  A dependency parse
for $\boldsymbol{x}$, denoted $\boldsymbol{y}$, is a set of tuples $(h, m, \ell)$, where $h$ is the index of a head, $m$ the index of a modifier, and $\ell$ a dependency label (or relation type).  Most dependency parsers are constrained to return $\boldsymbol{y}$ that form a directed tree.  

A first-order graph-based (\textbf{FOG}; also known as ``arc-factored'') dependency parser exactly solves
\begin{equation}
\boldsymbol{\hat{y}}(\boldsymbol{x}) = \arg\max_{\boldsymbol{y} \in \mathcal{T}(\boldsymbol{x})} \underbrace{\sum_{(h, m) \in \boldsymbol{y}} s(h, m, \boldsymbol{x})}_{S(\boldsymbol{y}, \boldsymbol{x})},
\label{eq:fog}
\end{equation}
where $\mathcal{T}(\boldsymbol{x})$ is the set of directed trees over $\boldsymbol{x}$, and
$s$ is a local scoring function that considers only a single dependency arc at a time.  (We 
suppress dependency labels; there are various ways to incorporate them, discussed later.)
To define $s$, \newcite{online_large_margin} used hand-engineered features of the surrounding and in-between context of $x_h$ and $x_m$; more recently, \newcite{kiperwasser} used a bidirectional LSTM followed by a single hidden layer with non-linearity.

The exact solution to Eq.~\ref{eq:fog} can be found using a minimum (directed) spanning tree algorithm \cite{nonprojective_mcdonald} 
 or, under a projectivity constraint, a dynamic programming algorithm \cite{eisner}, in $O(n^2)$ or $O(n^3)$ runtime, respectively. We refer to parsing with a minimum spanning tree algorithm as \textbf{MST parsing}.

An alternative that runs in linear time is \textbf{transition-based} parsing, which recasts parsing as a sequence of actions that manipulate auxiliary data structures to incrementally build a parse tree \cite{Nivre2003}. Such parsers can return a solution in a faster $O(n)$ asymptotic runtime. Unlike FOG parsers, transition-based parsers allow the use of scoring functions with history-based features, so that attachment decisions can interact more freely; the best performing parser at the time of this writing employ neural networks \cite{globally_normalized}.
\ignore{\miguelcomment{The idea of adding the paragraph above as a footnote was if the below paragraph was included... not sure like this (the paragraph with the cheap alternative is ignored), but probably as it is now. Be careful with the claim about richer than FOG parsers... this might "offend" people and we do not want that.}}

\ignore{Another cheap alternative, which might result in a parse that is not a well-formed tree, is to construct $\boldsymbol{\hat{y}}(\boldsymbol{x})$ as
$\{(\arg\max_h s(h,m,\boldsymbol{x}), m) \}_{m=1}^n$, i.e., greedily assign each word its highest-scoring head.  This is a relaxation of Eq.~\ref{eq:fog}, hence we call it \textbf{relaxed}  parsing.  A na{\"i}ve implementation runs in $O(n^2)$ time\footnote{The quadratic complexity arises from looping over all head-modifier attachment pairs}, although without the guarantee of finding a well-formed tree. \ignore{\nascomment{Actually, I'm not sure it does.  There are $n$ words, and we need to find the max for each one.  So this might be better than quadratic if you're clever, but naively it's quadratic.  so maybe drop that last sentence.}\cjd{I think this is wrong. The obvious algorithm is to max over each word at each position, so $n times n$. When counting votes, you might be able to use a heap or something and get something like $n \log k$ where $k$ is the number of things in the ensemble, but that's a bit too fussy. I'd just say this is quadratic.} \lingpengcomment{I think it's quadratic} \miguelcomment{I agree.}}}

Let $h_{\boldsymbol{y}}(m)$ denote the parent of $x_m$ in $\boldsymbol{y}$ (using a special null symbol when $m$ is the root of the tree), and $h_{\boldsymbol{y'}}(m)$ denotes the parent of $x_m$ in the predicted tree $\boldsymbol{y'}$. Given two dependency parses of the same sentence, $\boldsymbol{y}$ and $\boldsymbol{y'}$, the \textbf{Hamming cost}  is 
\begin{align}
C_H(\boldsymbol{y}, \boldsymbol{y'}) 
& = \sum_{m=1}^n \left\{ \begin{array}{ll} 0 & \mbox{if $h_{\boldsymbol{y}}(m) = h_{\boldsymbol{y'}}(m) $} \\ 1 & \mbox{otherwise} \end{array} \right. \nonumber
\end{align}
\ignore{\lingpengcomment{I don't quite like the expension of the equation... took me a while to understand...the $h : (h, m) \in \boldsymbol{y} \cap \boldsymbol{y'}$ notation is confusing. maybe use $h_{\boldsymbol{y}}(m))$ and indiction function?}\adhicomment{Addressed}}This cost underlies the standard dependency parsing evaluation scores (unlabeled and labeled attachment scores, henceforth UAS and LAS).  More generally, a \textbf{cost function} $C$ maps pairs of parses for the same sentence to non-negative values interpreted as the cost of mistaking one for the other, and a \textbf{first-order cost function} (FOC) is one that decomposes by attachments, like the Hamming cost.

Given a cost function $C$ and a probabilistic model that defines $p(\boldsymbol{y} \mid \boldsymbol{x})$, \textbf{minimum Bayes risk} (MBR) decoding is defined by
\begin{align}
\boldsymbol{\hat{y}}_{\text{MBR}}(\boldsymbol{x}) & = \arg\min_{\boldsymbol{y} \in \mathcal{T}(\boldsymbol{x})} \sum_{\boldsymbol{y'} \in \mathcal{T}(\boldsymbol{x})} p(\boldsymbol{y'} \mid \boldsymbol{x}) \cdot C(\boldsymbol{y}, \boldsymbol{y'}) \nonumber \\
& =
\arg \min_{\boldsymbol{y} \in \mathcal{T}(\boldsymbol{x})}  \mathbb{E}_{p(\boldsymbol{Y} \mid \boldsymbol{x})}[C(\boldsymbol{y}, \boldsymbol{Y})].
\label{eq:mbr}
\end{align}
Under the Hamming cost, MBR parsing equates algorithmically to FOG parsing with $s(h, m, \boldsymbol{x}) = p((h, m) \in \boldsymbol{Y} \mid \boldsymbol{x})$, the posterior marginal of the attachment under $p$.  This is shown by linearity of expectation; see also \newcite{titov:2006}.

Apart from MBR decoding, cost functions are also used for \emph{discriminative training} \ignore{\adhicomment{can we say that this is a non-probabilistic use of cost?} use of cost functions is in \textbf{discriminative training} \adhicomment{R2 says that this part is confusing and that we should elaborate more what we mean by 'a rather different use of cost function'}}of a parser.  For example, suppose we seek to estimate the parameters $\boldsymbol\theta$ of scoring function $S_{\boldsymbol{\theta}}$.  One approach is to minimize the structured hinge loss of a training dataset $\mathcal{D}$ with respect to $\boldsymbol{\theta}$:

\begin{align}
\min_{\boldsymbol{\theta}} \sum_{(\boldsymbol{x}, \boldsymbol{y}) \in \mathcal{D}} [ & -\ S_{\boldsymbol{\theta}}(\boldsymbol{y}, \boldsymbol{x}) \nonumber \\
 & +\ \max_{\boldsymbol{y'} \in \mathcal{T}(\boldsymbol{x})}\left(  S_{\boldsymbol{\theta}}(\boldsymbol{y'}, \boldsymbol{x}) + C(\boldsymbol{y'}, \boldsymbol{y})\right)]
 \label{eq:hinge}
 \end{align}
\ignore{\lingpengcomment{I put the neg first... I think this way is clear in format, if it is impossible to put everything in a single line.}}
Intuitively, this amounts to finding parameters that separate the model score of the correct parse from any wrong parse by a distance proportional to the cost of the wrong parse.
With regularization, this is equivalent to the structured support vector machine \cite{taskar,tsochantaridis}, \ignore{\nascomment{missing cites}}and if $S_{\boldsymbol{\theta}}$ is (sub)differentiable, many algorithms are available.
Variants have been used extensively in training graph-based parsers \cite{nonprojective_mcdonald,martins-09}, which typically
make use of Hamming cost, so that the inner $\max$ can be solved efficiently using FOG parsing with a slightly revised local scoring function:
\begin{equation}
s'(h, m, \boldsymbol{x})  = s(h, m, \boldsymbol{x}) + \left\{ \begin{array}{ll} 0 & \mbox{if $(h,m) \in \boldsymbol{y}$} \\ 1 & \mbox{otherwise} \end{array} \right.
\end{equation}
Plugging this into Eq.~\ref{eq:fog} is known as \textbf{cost-augmented} parsing.

\section{Consensus and Minimum Bayes Risk}\label{sec:mbr}

Despite the recent success of neural network dependency parsers, most prior works exclusively report single-model performance. Ensembling neural network models trained from different random starting points is a standard technique in a variety of problems, such as machine translation \cite{seq_to_seq} and constituency parsing \cite{grammar_foreign}. We aim to investigate the benefit of ensembling independently trained neural network dependency parsers by applying the parser ensembling method of \newcite{ensemble_reparsing} to a collection of $N$ strong neural network base parsers. \adhicomment{Added some sentences as a more gentle introduction to ensembling}

Here, each base parser is an instance of the greedy, transition-based parser of \newcite{stack_lstm}, known as the stack LSTM parser, trained from a different random initial estimate.  Given a sentence $\boldsymbol{x}$, the consensus FOG parser (Eq.~\ref{eq:fog}) defines  score $s(h, m, \boldsymbol{x})$ as the number
of base parsers that include the attachment $(h, m)$, which we denote $\mathit{votes}(h,m)$.\footnote{An alternative to building an ensemble of stack LSTM parsers in this way would be to average the softmax decisions at each timestep (transition), similar to \newcite{grammar_foreign}. \ignore{\cjd{they didn't do dep parsing, but they did it in this paper}  Our approach is simpler to implement and more memory-efficient, since each parser can be run sequentially.  Further, voting on each attachment is more amenable to the distillation we will explore in \S\ref{sec:distillation}. \nascomment{Chris wanted us to make this clear; I think it would be good to cite the kind of ensemble he suggests is an alternative, but I'm not sure what the citation would be}\adhicomment{I'm not aware of any works that did that.}}} An example of this scoring function with an ensemble of 20 models is shown in Figure~\ref{ensembling_figure} We assign to dependency $(h, m)$ the label most frequently selected by the base parsers that attach $m$ to $h$.

\begin{figure}
\begin{dependency}
  \centering
  \begin{deptext}
  John \& \textbf{saw} \& the \& \textbf{woman} \& \emph{with} \& a \& telescope\\
  \end{deptext}
  \depedge{5}{2}{19}
  \depedge{5}{4}{1}
     \end{dependency}
     \caption{Our ensemble's votes (20 models) on an ambiguous PP attachment of \emph{with}. \ignore{\nascomment{I took out the 0 to John; why show that when there's also 0 to the, with, a, and telescope? also switched arrow direction}}The ensemble is nearly but not perfectly unanimous in selecting \emph{saw} as the head.}
     \label{ensembling_figure}
 \end{figure}

Next, note that if we let $s(h,m, \boldsymbol{x}) = \mathit{votes}(h,m)/N$, this has no effect on the parser (we have only scaled by a constant factor).  We can therefore view $s$ as a posterior marginal, and the ensemble parser as an MBR parser (Eq.~\ref{eq:mbr}).   


\paragraph{Experiment.}

We consider this approach on the Stanford dependencies version 3.3.0 \cite{stanford_dependencies} Penn Treebank task.
As noted, the base parsers instantiate the greedy stack LSTM parser \cite{stack_lstm}.\footnote{We use the standard data split (02--21 for training, 22 for development, 23 for test), automatically predicted part-of-speech tags, same pretrained word embedding as \newcite{stack_lstm}, and recommended hyperparameters; \url{https://github.com/clab/lstm-parser},\ignore{\miguelcomment{and pretrained word embeddings -> just say that it is the same setting as the best result of dyer et al. 2015}} each with a different random initialization; this differs
from past work on ensembles, which often uses different base model architectures. \ignore{\miguelcomment{We are using different base models since they have different random initialization... maybe explain (accuracy)? otherwise the reader might be confused with the fact that ensembling the same parser 20 times improves the accuracy}}  \ignore{\nascomment{need to explain jackknifing here?}}}


\begin{table}
\centering
\resizebox{0.95\columnwidth}{!}{%
\begin{tabular}{|l|r|r|r|}
\hline
\multicolumn{1}{|c|}{\textbf{Model}} & \multicolumn{1}{c|}{\textbf{UAS}} & \multicolumn{1}{c|}{\textbf{LAS}} & \multicolumn{1}{c|}{\textbf{UEM}}\\ \hline
\newcite{globally_normalized} & \textbf{94.61} & \textbf{92.79} & - \\ 
$N =1$ (stack LSTM)                   & 93.10                              & 90.90   & 47.60                           \\ 

ensemble, $N = 5$, MST            & 93.91                             & 91.94    & 50.12                         \\ 
ensemble, $N = 10$, MST             & 94.34                             & 92.47    & 52.07                        \\ 
ensemble, $N= 15$, MST             & 94.40                             & 92.57    & 51.86                         \\ 
ensemble, $N = 20$, MST            & 94.51                             & 92.70    & \textbf{52.44}                         \\ \hline
\end{tabular}
}
\caption{PTB-SD task:  ensembles improve over a strong greedy baseline.
UEM indicates unlabeled exact match. \ignore{\nascomment{don't forget to fill in UEM numbers if we have them}}
\label{ensemble_result}}
\end{table}

Table~\ref{ensemble_result} shows that ensembles, even with small $N$,  strongly outperform a single stack LSTM parser.\ignore{This even holds with relaxed parsing, which preserves the $O(n)$ runtime guarantee and does not suffer on UAS or LAS.\cjd{this is not correct, see above}\lingpengcomment{it feels kind of meaningless to talk about UAS in relatxed parsing... UAS is indeed measurable in relaxed parsing, but since it give up the well-formal tree constraint, does not suffer on UAS doesn't seem to be a real advantage? personally view though... Is it really necessary to talk about relax parsing?}}  Our ensembles of greedy, locally normalized parsers perform comparably to the best previously reported, due to \newcite{globally_normalized}, which uses a beam (width 32) \nascomment{please check} for training and decoding. 

\section{What is Ensemble Uncertainty?}\label{sec:ensemble_ambiguity}

While previous works have already demonstrated the merit of ensembling in dependency parsing \cite{ensemble_reparsing,ensemble_cheap_and_good}, usually with diverse base parsers, we consider whether the posterior
marginals estimated by $\hat{p}((h, m) \in \boldsymbol{Y} \mid \boldsymbol{x}) = \mathit{votes}(h, m)/N$ can be interpreted.  
We conjecture that disagreement among base parsers about where to attach $x_m$ (i.e., uncertainty in the posterior) is a sign of difficulty or ambiguity.  If this is true, then the ensemble provides information about which confusions are more or less reasonable---information we will exploit in our distilled  parser (\S\ref{sec:distillation}).

A complete linguistic study is out of scope here; instead, we provide a motivating example before empirically validating our conjecture.
Table~\ref{tab:qualitative_analysis} shows an example where there is considerable disagreement among base parsers over the attachment of a word (\emph{including}). We invite the reader to attempt to select the correct attachment and gauge the difficulty of doing so, before reading on.

Regardless of whether our intuition that this is an inherently difficult and perhaps ambiguous case is correct, it is uncontroversial to say that the words in the sentence not listed, which received zero votes (e.g., both instances of \emph{the}), are obviously implausible attachments.

\begin{table}[t]
\ignore{\miguelcomment{Add Sentence: here or Example: here}} \textbf{Sentence:} It will \textbf{go} for \textbf{work} ranging from refinery\\ \textbf{modification} to \textbf{changes} in the distribution \textbf{system},\\ \emph{including} the way service stations \textbf{pump} fuel into cars.

\begin{center}
\begin{tabular}{|c|r|r|r|}
\hline
\multicolumn{1}{|c}{$x_h$}                                                                       & \multicolumn{1}{|c|}{posterior}        &
\multicolumn{1}{c|}{new cost}                                                           &
\multicolumn{1}{c|}{Hamming}                                                               \\ \hline
go                                                                                               & 0.143   &  0.143    & 1                                                                                                            \\
work                                                                                                & 0.095  & 0.191 & 1                                                                                                                 \\
modification                                                                                        & 0.190            & 0.096     &1                                                                                                    \\
changes                                                                                             & \textbf{0.286}   & \textbf{0.000} & \textbf{0}                                                                                                                  \\
system                                                                                              & 0.095     & 0.191    & 1                                                                                                       \\
pump                                                                                               & 0.190 &0.096 & 1 \\
stations & 0.000 & 0.286 & 1 \\ \hline                                                                                                                   
\end{tabular}
\end{center}
\caption{An ambiguous sentence from the training set and the posteriors\protect\footnotemark{} of various possible parents for \emph{including}. 
The last two columns are, respectively, the contributions to the distillation cost $C_D$
(explained in \S\ref{sec:distillation-cost}, Eq.~\ref{eq3}) and the standard Hamming cost $C_H$. The most probable head under the ensemble is \emph{changes}, which is also the correct answer.} 
\label{tab:qualitative_analysis}
\end{table}
\footnotetext{\label{note8}In \S\ref{sec:mbr}, we used 20 models. Since those 20 models were trained on the whole training set, they cannot be used to obtain the uncertainty estimates on the training set, where the example sentence in Table \ref{tab:qualitative_analysis} comes from. Therefore we trained a new ensemble of 21 models from scratch with five-way jackknifing. The same jackknifing setting is used in the distillation parser (\S\ref{sec:exp}).}

Our next idea is to transform ensemble uncertainty into a new estimate of cost---a replacement for the Hamming cost---and use it in discriminative training of a single FOG parser.  This allows us to distill what has been learned by the ensemble into a single model.




\section{Distilling the Ensemble} \label{sec:distillation}
Despite its state of the art performance, our ensemble requires $N$ parsing calls to decode each sentence.  To reduce the computational cost, we introduce a method for
``distilling'' the ensemble's knowledge into a single parser, making use of a novel \textbf{cost function} to communicate this knowledge from the ensemble to the distilled model.
While models that combine the outputs of other parsing models have been proposed before \cite[\emph{inter alia}]{stacking,NivreIGA08,combine_beam_search}, these works incorporated the scores or outputs of the baseline parsers as \textbf{features} and as such require running the first-stage models at test-time.\cjd{there was way too much detail that was interrupting the exposition of the new idea here, so i cut some stuff} 
Creating a cost function from a data analysis procedure is, to our knowledge, a new idea. 

The idea is attractive because cost functions are model-agnostic; they can be used with any parser amenable to discriminative training.  Further, only the training procedure changes; parsing at test time does not require consulting the ensemble at all, avoiding the costly application of the $N$ parsers to new data, unlike model combination techniques like stacking and beam search.

Distilling an ensemble of classifiers into one simpler classifer that behaves similarly is due to \newcite{bucila:2006} and \newcite{dark_knowledge}; they were likewise motivated by a desire to create a simpler model that was cheaper to run at test time.  In their work, the ensemble provides a probability distribution over labels for each input, and this predicted distribution serves as the training target for the distilled model (a sum of two cross entropies objective is used, one targeting the empirical training distribution and the other targeting the ensemble's posterior distribution).
This can be contrasted with the supervision provided by the training data alone, which conventionally provides a single correct label for each instance.  These are respectively called ``soft'' and ``hard'' targets.

\begin{table}[!ht]
\ignore{\miguelcomment{Same here}}
 \textbf{Sentence: }John \textbf{saw} the \textbf{woman} \emph{with} a telescope
\begin{center} \begin{tabular}{|c|r|r|} \hline
$x_h$                                & soft  & hard                        \\ \hline
John                                              & 0.0 & 0 \\ 
 saw                                               & \textbf{0.95} &  \textbf{1}                                                   \\
the & 0.0 & 0 \\
woman                                             & 0.05  &  0                                                 \\
a & 0.0 & 0 \\
telescope & 0.0 & 0 \\
\hline                                                
\end{tabular}
\end{center}
\caption{Example of soft targets (taken from \ignore{\nascomment{check that this is actually what our model does, or fix}} our 20-model ensemble's uncertainty on the sentence) and hard targets (taken from the gold standard) for possible parents of \emph{with}. The soft target corresponds with the posterior (second column) in Table \ref{tab:qualitative_analysis}, but the hard target differs from the Hamming cost (last column of Table \ref{tab:qualitative_analysis}) since the hard target assigns a value of 1 to the correct answer and 0 to all others (the reverse is true for Hamming cost).
\label{tab:soft_hard_targets}}
\end{table}


We propose a novel adaptation of the soft target idea to the structured output case. Since a sentence has an exponential (in its length) number of parses, representing the posterior distribution over parses predicted by the ensemble is nontrivial. We solve this problem by taking a single parse from each model, representing the $N$-sized ensemble's parse distribution using $N$ samples.

Second, rather than considering uncertainty at the level of complete parse trees (which would be analogous to the classification case) or larger structures, we instead consider uncertainty about \emph{individual} attachments, and seek to ``soften'' the attachment targets used in training the parser. An illustration for the prepositional phrase attachment ambiguity in Fig.~\ref{ensembling_figure}, taken from the ensemble output for the sentence, is shown in Table~\ref{tab:soft_hard_targets}. Soft targets allow us to encode the notion that mistaking \emph{woman} as the parent of \emph{with} is less bad than attaching \emph{with} to \emph{John} or \emph{telescope}.  Hard targets alone do not capture this information.


\subsection{Distillation Cost Function} \label{sec:distillation-cost}

The natural place to exploit this additional information when training a parser is in the cost function. When incorporated into discriminative training, the Hamming cost encodes hard targets:  the correct attachment should receive
a higher score than all incorrect ones, with the same margin.  Our distillation cost function aims to reduce the cost of decisions
that---based on the ensemble uncertainty---appear to be more difficult, or where there may be multiple plausible attachments.

Let $\pi(h, m) = $
\begin{equation*}
  1 - \hat{p}((h, m) \in \boldsymbol{Y} \mid \boldsymbol{x}) =  \frac{N- \mathit{votes}(h, m)}{N}.
\end{equation*}
Our new cost function is defined by $C_D(\boldsymbol{y}, \boldsymbol{y'}) =$
\begin{align}
\textstyle \sum_{m=1}^n \max \left\{ 0, \pi(h_{\boldsymbol{y'}}(m), m) - \pi(h_{\boldsymbol{y}}(m), m) \right\} \nonumber \\
= \textstyle \sum_{m=1}^n \max \left\{0, \hat{p}(h_{\boldsymbol{y}}(m),m) - \hat{p}(h_{\boldsymbol{y'}}(m),m) \right\}. \label{eq3} 
\end{align}

Recall that $\boldsymbol{y}$ denotes the correct parse, according to the training data, and $\boldsymbol{y'}$ is a candidate parse.

This function has several attractive properties:
\begin{enumerate}
\item When a word $x_m$ has more than one plausible (according to the ensemble) but incorrect (according to the annotations) attachment, each one has a diminished cost (relative to Hamming cost and all implausible attachments).
\item The correct attachment (according to the gold-standard training data) always has zero cost since $h_{\boldsymbol{y}}(m)=h_{\boldsymbol{y'}}(m)$ and Eq.~\ref{eq3} cancels out. \ignore{\nascomment{I may have misunderstood what you wrote}}
\item When the ensemble is confident, cost for its choice(s) is lower than it would be under Hamming cost---even when the ensemble is wrong.  This means that we are largely training the distilled parser to simulate the ensemble, including mistakes and correct predictions. This encourages the model to replicate the state of the art ensemble performance.
\item Further, when the ensemble is perfectly confident and correct, every incorrect attachment  has a cost of 1, just as in Hamming cost.
\item The cost of any attachment is bounded above by the proportion of votes assigned to the correct attachment.  \ignore{\nascomment{check}}
\end{enumerate}


One way to understand this cost function is to imagine that it gives the parser more ways to achieve a zero-cost\footnote{It is important to note the difference between cost (Eq.~\ref{eq3}) and loss (Eq.~\ref{eq:hinge}).} attachment.  The first is to correctly attach a word to its correct parent.  The second is to predict a parent that the ensemble prefers to the correct parent, i.e.,
$\pi(h_{\boldsymbol{y'}}(m), m) < \pi(h_{\boldsymbol{y}}(m), m)$. Any other decision will incur a non-zero cost that is proportional to the implausibility of the attachment, according to the ensemble. Hence the model is supervised both by the hard targets in the training data annotations and the soft targets from the ensemble.

While it may seem counter-intuitive to place zero cost on an incorrect attachment, recall that the \emph{cost} is merely a margin that must separate the scores of parses containing correct and incorrect arcs. In contrast, the \emph{loss} (in our case, the structured hinge loss) is the “penalty” the learner tries to minimize while training the graph-based parser, which depends on both the \emph{cost} and \emph{model score} as defined in Equation 3. When an incorrect arc is preferred by the ensemble over the gold arc (hence assigned a cost/margin of 0), the model will still incur a loss if $s(h_{\boldsymbol{y}}(m), m, \boldsymbol{x}) < s(h_{\boldsymbol{y'}}(m), m, \boldsymbol{x})$. In other words, the score of \emph{any} incorrect arc (including those strongly preferred by the ensemble) cannot be higher than the score of the gold arc. 

The learner only incurs 0 loss if $s(h_{\boldsymbol{y}}(m), m, \boldsymbol{x}) \geq s(h_{\boldsymbol{y'}}(m), m, \boldsymbol{x})$. This means that the gold score and the predicted score can have a margin of 0 (i.e., have the same score and incur no loss) when the ensemble is highly confident of that prediction, but the score of the correct parse cannot be lower regardless of how confident the ensemble is (hence the objective does not encourage incorrect trees at the expense of gold ones).

In the example in Table~\ref{tab:qualitative_analysis}, we show the (additive) contribution to the distillation cost by each attachment decision (column labeled ``new cost'').  Note that more plausible attachments according to the ensemble have a lower cost than less plausible ones (e.g., the cost for \emph{modification} is less than \emph{system}, though both are incorrect). While in the last line \emph{stations} received no votes in the ensemble (implausible attachment), its contribution to the cost is bounded by the proportion of votes for correct attachment. The intuition is that, when the ensemble is not certain of the correct answer, it should not assign a large cost to implausible attachments. In contrast, Hamming cost would assign a cost of 1 (column labeled ``Hamming'') in all incorrect cases.

\subsection{Distilled Parser}
Our distilled parser is trained  discriminatively with the structured hinge loss (Eq.~\ref{eq:hinge}).  This is a natural choice because it makes the cost function explicit and central to learning.\footnote{Alternatives that do not use cost functions include probabilistic parsers, whether locally normalized like the stack LSTM parser used within our ensemble, or globally normalized, as in \newcite{globally_normalized}; cost functions can be incorporated in such cases with minimum risk training \cite{smith-eisner:2006:POS} or softmax margin \cite{gimpel2010softmax}.}  Further, because our ensemble's posterior
gives us information about each attachment individually, the cost function we construct can be first-order, which simplifies training with exact inference.  

This approach to training a model is well-studied for a FOG parser, but not for a  transition-based parser, which is comprised of a collection of classifiers trained to choose good sequences of transitions---not to score whole trees for good attachment accuracy. Transition-based approaches are therefore unsuitable for our proposed distillation cost function, even though they are asymptotically faster. \ignore{\nascomment{might want to discuss more here.}} \ignore{\miguelcomment{why do we relate to transntion-based parsers here?}}We proceed with a FOG parser (with Eisner's algorithm for English and Chinese, and MST for German since it contains a considerable number of non-projective trees)\ignore{\miguelcomment{explain why mst for german and eisner's for the rest...}} as the distilled model.

Concretely, we use a bidirectional LSTM followed by a hidden layer of non-linearity to calculate the scoring function $s(h, m, \boldsymbol{x})$, following \newcite{kiperwasser} with minor modifications. 
The bidirectional LSTM maps each word $x_i$ to a vector $\mathbf{\bar{x}}_i$ that embeds the word in context (i.e., $\boldsymbol{x}_{1:i-1}$ and $\boldsymbol{x}_{i+1:n}$).  Local attachment scores are given by:
\begin{equation}
s(h, m, \boldsymbol{x}) = \mathbf{v}^\top \tanh\left ( \mathbf{W} [\mathbf{\bar{x}}_h ; \mathbf{\bar{x}}_m ] + \mathbf{b} \right)
\end{equation}
where the model parameters are $\mathbf{v}$, $\mathbf{W}$, and $\mathbf{b}$, plus the bidirectional LSTM parameters.
 We will refer to this parsing model as \textbf{neural FOG}.

Our model architecture is nearly identical to that of \newcite{kiperwasser}, with two primary differences. The first difference is that we fix the pretrained word embeddings and compose them with  learned embeddings and  POS tag embeddings \cite{stack_lstm}, allowing the model to simultaneously leverage pretrained vectors and learn a task-specific representation.\footnote{To our understanding, \newcite{kiperwasser} initialized with pretrained vectors and backpropagated during training.} Unlike \newcite{kiperwasser}, we did not observe any degradation by incorporating the pretrained vectors. Second, we apply a per-epoch learning rate decay of 0.05 to the Adam optimizer. While the Adam optimizer automatically adjusts the global learning rate according to past gradient magnitudes, we find that this additional per-epoch decay consistently improves performance across all settings and languages.  \ignore{First, we concatenate pretrained word embeddings (kept fixed during training) with word embeddings learned from scratch and part-of-speech tag embeddings, followed by a rectifier non-linearity, to obtain the word representation fed into the bidirectional LSTM \cite{stack_lstm}. This approach allows the model to use pretrained vectors while simultaneously learning a representation suitable for parsing.  We understand \newcite{kiperwasser} to have used the pretrained word embedding to initialize the word vectors and backpropagate during training. Unlike \newcite{kiperwasser}, we did not observe any degradation in performance when incorporating pretrained word embeddings to our model. Second, we incorporate a per-epoch decay to the Adam optimizer \cite{adam}, which is beneficial to model performance under all settings.  \nascomment{what did Kiperwaser do (why is this framed as a second difference?  don't assume the reader knows every detail of that paper}\cjd{the details about word embeddings seem a bit complicated. i think we should just be precise in what we did, and say that this is very close to their model}\lingpengcomment{second thing we can juse say in experiment section we use adam, and the first thing seems not that important to me too. this is the way the stack-lstm do the input right? probably just say we construct the input same as that? say we build a model very close to kiperwasser is probably enough?}} 


\section{Experiments}\label{sec:exp}
We ran experiments on the English PTB-SD version 3.3.0, Penn Chinese Treebank \cite{ctb}, and German CoNLL 2009 \cite{conll2009} tasks. \ignore{\miguelcomment{why do you cite ptb and conll, but you don't the ctb treebank? you already cited PTB-SD above}}

\textbf{Experimental settings}. We used the standard splits for all languages.  Like \newcite{chen_manning} and \newcite{stack_lstm}, we use predicted tags with the Stanford tagger \cite{stanford_tagger} for English and gold tags for Chinese. For German we use the predicted tags provided by the CoNLL 2009 shared task organizers. All models were augmented with pretrained structured-skipgram \ignore{\nascomment{better to call this ``structured-skipgram''?  since w2v is a package, not a model}}\cite{wang2vec} embeddings;
for English we used the Gigaword corpus and 100 dimensions, for Chinese Gigaword and 80, and for German WMT 2010 monolingual data and 64.

\textbf{Hyperparameters}. The hyperparameters for neural FOG are summarized in Table~\ref{hyperparameters}. For the Adam optimizer we use the default settings in the CNN neural network library.\footnote{\url{https://github.com/clab/cnn.git}} Since the ensemble is used to obtain the uncertainty on the training set, it is imperative that the stack LSTMs do not overfit the training set. To address this issue, we performed  five-way jackknifing of the training data for each stack LSTM model to obtain the training data uncertainty under the ensemble.\ignore{\nascomment{this doesn't make sense ... uncertainty for each stack LSTM? huh?}} To obtain the ensemble uncertainty on each language, we use 21 base models for English (see footnote \ref{note8}),\ignore{\lingpengcomment{here is 21 again, the intro is 20}\adhicomment{This is because the ensemble and the ensemble-jackknifing models are different. That's why we don't have results for Chinese and German.}\miguelcomment{...the reviewers are going to wonder the same, we need to explain it.}\adhicomment{Addressed with a footnote.}} 17 for Chinese, and 11 for German.\ignore{\cjd{I agree with Noah's objection. Maybe say something like since we want to decode the training data with the ensemble, we use 5-way jackknifing to avoid overfitting the transition-based parsers in the ensemble.}\adhicomment{Addressed}} 
\begin{table}[]
\centering
\resizebox{0.7\columnwidth}{!}{%
\begin{tabular}{|l|l|}
  \hline
Bi-LSTM dimension                                                        & 100                                                                  \\ 
Bi-LSTM layers                                                             & 2                                                                    \\ 
POS tag embedding                                                          & 12                                                                   \\ 
Learned word embedding         & 32                                                                   \\ 
Hidden Layer Units                                                                       & 100                                                                  \\ 
Labeler Hiden Layer Units                                                               & 100                                                                  \\ 
Optimizer                                                                  & Adam                                                                \\ 
Learning rate decay    & 0.05                                                                 \\ \hline
\end{tabular}
}
\caption{Hyperparameters for the distilled FOG parser. Both the model architecture and the hyperparameters are nearly identical with Kiperwasser and Goldberg (2016). We apply a per-epoch learning rate decay to the Adam optimizer, which consistently improves performance across all datasets. \ignore{MLP indicates a one hidden layer multi-layer perceptron after \nascomment{what is ``MLP''?}} 
\label{hyperparameters}}
\end{table}

\begin{table*}[!ht]
\centering
  \resizebox{2\columnwidth}{!}{%
  \begin{tabular}{|l|l|c|l|l|l|l|l|l|}
  \hline
\multirow{2}{*}{\textbf{System}} & \multirow{2}{*}{\textbf{Method}} & \multirow{2}{*}{\textbf{\begin{tabular}[c]{@{}c@{}}P? \end{tabular}}} & \multicolumn{2}{c|}{\textbf{PTB-SD}} & \multicolumn{2}{c|}{\textbf{CTB}} & \multicolumn{2}{c|}{\textbf{\begin{tabular}[c]{@{}c@{}}German\\ CoNLL'09\end{tabular}}} \\ \cline{4-9}
                                 &                                  &                                                                                            & \textbf{UAS}      & \textbf{LAS}     & \textbf{UAS}    & \textbf{LAS}    & \textbf{UAS}                               & \textbf{LAS}                               \\ \hline
\newcite{zhang_nivre_11}                    & Transition (beam)                &                                                                                                    & -                 & -                & 86.0            & 84.4            & -                                           & -                                           \\
\newcite{bohnet_nivre_12}$^{\dagger}$                    & Transition (beam)                &                                                                                                     & -                 & -                & 87.3            & 85.9            & 91.37                                           & \underline{89.38}                                           \\
\newcite{chen_manning}                    & Transition (greedy)              & $\checkmark$                                                                                                    & 91.8              & 89.6             & 83.9            & 82.4            & -                                           & -                                           \\
\newcite{stack_lstm}                           & Transition (greedy)              & $\checkmark$                                                                                                     & 93.1              & 90.9             & 87.2            & 85.7            & -                                           & -                                           \\
\newcite{structured_training}                          & Transition (beam)                & $\checkmark$                                                                                                    & 94.0              & 92.0             & -               & -               & -                                            & -                                           \\
\newcite{yazdani_henderson_15}                          & Transition (beam)                &                                                                                                     & -              & -            & -               & -               & 89.6                                            & 86.0                                           \\
\newcite{ballesteros_et_al_15}                          & Transition (greedy)                &                                                                                                       & 91.63              & 89.44             & 85.30               & 83.72               & 88.83                                            & 86.10                                           \\
\newcite{dynamic_oracle}                    & Transition (greedy)             & $\checkmark$                                                                                                & 93.56             & 91.42            & 87.65           & 86.21           & -                                            & -                                           \\
\newcite{kiperwasser}           & Transition (greedy)             & $\checkmark$                                                                                                      & 93.9              & 91.9             & 87.6            & 86.1            &  -                                          & -                                           \\
\newcite{globally_normalized}                          & Transition (beam)               & $\checkmark$                                                                                                     & \textbf{\underline{94.61}}    & \textbf{\underline{92.79}}   & -               & -               & 90.91                                            & 89.15                                           \\ \hline
\newcite{fourth_order}                         & Graph (4th order)                &                                                                                                   & -                 & -                & 87.74           & -               & -                                           & -                                           \\
\newcite{turbo_parser}                        & Graph (3rd order)               &                                                                                                 & 93.1              & -                & -               & -               &  -                                          &  -                                          \\
\newcite{le_zuidema}                      & Reranking/blend                 & $\checkmark$                                                                                               & 93.8              & 91.5             & -               & -               & -                                           & -                                           \\
\newcite{zhu_et_al_15}                           & Reranking/blend                 & $\checkmark$                                                                                             & -                 & -                & 85.7            & -               & -                                           & -                                           \\
\newcite{kiperwasser}            & Graph (1st order)                &                                                                                                  & 93.1              & 91.0             & 86.6            & 85.1            & -                                           & -                                           \\
\newcite{graph_based_segment}                           & Graph (1st order)                 & $\checkmark$                                                                                             & 94.08                 & 91.82                & 87.55            & 86.23               & -                                           & -                                           \\
 \hline \hline
This work:  ensemble, $N=20$, MST   & Transition (greedy)     &   $\checkmark$      & 94.51                    & 92.70 &\textbf{89.80} &\textbf{88.56} &\textbf{91.86} &\textbf{89.98} \\ 
This work:  neural FOG, $C_H$          & Graph (1st order)                & $\checkmark$                                                                                           & 93.76             & 91.60            & 87.32           & 85.82           & 91.22                                           &  88.82                                          \\
This work:  neural FOG, $C_D$ (distilled)     & Graph (1st order)                & $\checkmark$                                                                                     & 94.26             & 92.06            & \underline{88.87}  & \underline{87.30}  & \underline{91.60}                                           & 89.24  \\ \hline                                         
\end{tabular}
}
  \caption{Dependency parsing performance on English, Chinese, and German tasks.  The ``P?'' column indicates the use of pretrained word embeddings. Reranking/blend indicates that the reranker score is interpolated with the base model's score. Note that previous works might use different predicted tags for English. We report accuracy without punctuation for English and Chinese, and with punctuation for German, using the standard evaluation script in each case. We only consider systems that do not use additional training data. The best overall results are indicated with bold (this was achieved by the ensemble of greedy stack LSTMs in Chinese and German), while the best non-ensemble model is denoted with an underline. The $\dagger$ sign indicates the use of predicted tags for Chinese in the original publication, although we report accuracy using gold Chinese tags based on private correspondence with the authors. \ignore{$^\ast$Chinese results for these models are not directly comparable since they used predicted tags, while we follow more recent lines of work and use gold ones \protect \cite{stack_lstm,kiperwasser}.}\ignore{\miguelcomment{say who we follow better...}} \ignore{\miguelcomment{which ones? the relaxed ones? Not sure if I understand. If they are the relaxed ones, just leave the box empty, we don't have that numbers in the paper.}} \ignore{\nascomment{issue with the dagger?} \adhicomment{Resolved}
}}
  \label{tab:full_results}\ignore{\cjd{I think it's a bad idea to leave out the non-relaxed ensemble results}}
\end{table*}

\textbf{Speed.} One potential drawback of using a quadratic or cubic time parser to distill an ensemble of linear-time transition-based models is speed. 
Our FOG model is implemented using the same CNN library as the stack LSTM transition-based parser.
On the same single-thread CPU hardware, the distilled MST parser\footnote{The runtime of the Hamming-cost bidirectional LSTM FOG parser is the same as the distilled parser.}  parses 20 sentences per second without any pruning, while a single stack LSTM model is only three times faster at 60 sentences per second. Running an ensemble of 20 stack LSTMs is at least 20 times slower (without multi-threading), not including consensus parsing.
In the end, the distilled parser is more than ten times \ignore{\nascomment{I don't like that phrasing; just give a number like ``ten times"}}faster than the ensemble pipeline.



\textbf{Accuracy.}  All scores are shown in Table~\ref{tab:full_results}.
First, consider the neural FOG parser trained with Hamming cost ($C_{H}$ in the second-to-last row).  This is a very strong benchmark, \ignore{\miguelcomment{Difficult to call it "baseline" if it is the best result reported...}}outperforming many higher-order graph-based and neural network models on all three datasets.  Nonetheless, training the same model with distillation cost gives consistent improvements for all languages.
For English, we see that this model comes close to the slower ensemble it was trained to simulate.  For Chinese, it achieves the best published scores, for German the best published UAS scores, and just after \newcite{bohnet_nivre_12} for LAS.

\textbf{Effects of Pre-trained Word Embedding}. As an ablation study, we ran experiments on English without pre-trained word embedding, both with the Hamming and distillation costs. The model trained with Hamming cost achieved 93.1 UAS and 90.9 LAS, compared to 93.6 UAS and 91.1 LAS for the model with distillation cost. This result further showcases the consistent improvements from using the distillation cost across different settings and languages.

We conclude that ``soft targets'' derived from ensemble uncertainty offer useful guidance, through the distillation cost function and discriminative training
of a graph-based parser.  Here we considered a FOG parser, though future work might investigate any parser amenable to training to minimize a cost-aware  loss like the structured hinge.

\section{Related Work}
Our work on ensembling dependency parsers is based on \newcite{ensemble_reparsing} and \newcite{ensemble_cheap_and_good}; an additional contribution of this work is to show that the normalized\ignore{\lingpengcomment{normalized before and normalised here...}} ensemble votes correspond to MBR parsing. \newcite{petrov_2010} proposed a similar model combination with random initializations for phrase-structure parsing, using products of constituent marginals. The local optima in his base model's training objective arise from latent variables instead of neural networks (in our case).

Model distillation was proposed by \newcite{bucila:2006}, who used a single neural network to simulate a large ensemble of classifiers.  More recently, \newcite{ba_et_al_15} showed that a single {shallow} neural network can closely replicate the performance of an ensemble of {deep} neural networks in phoneme recognition and object detection. Our work is closer to \newcite{dark_knowledge}, in the sense that we do not simply \emph{compress} the  ensemble and hit the ``soft target,'' but also the ``hard target'' at the same time\footnote{Our cost is zero when the correct arc is predicted, regardless of what the soft target thinks, something a compression model without gold supervision cannot do.}. These previous works only used model compression and distillation for classification; we extend the work to a structured prediction problem (dependency parsing).

\newcite{tackstrom_et_al_13} similarly used an ensemble of other parsers to guide the prediction of a seed model, though in a different context of ``ambiguity-aware'' ensemble training to re-lexicalize a transfer model for a target language. We similarly use an ensemble of models as a supervision for a single model.  By incorporating the ensemble uncertainty estimates in the cost function, our approach is cheaper, not requiring any marginalization during training. An additional difference is that we learn from the gold labels (``hard targets'') rather than only ensemble estimates on unlabeled data. 

\newcite{seq-distillation} proposed a distillation model at the sequence level, with application in sequence-to-sequence neural machine translation. There are two primary differences with this work. First, we use a global model to distill the ensemble, instead of a sequential one. Second, \newcite{seq-distillation} aim to distill a larger model into a smaller one, while we propose to distill an ensemble instead of a single model.\adhicomment{Added new EMNLP paper from Sasha Rush's group}


\section{Conclusions}
We demonstrate that an ensemble of 20 greedy stack LSTMs \cite{stack_lstm} can achieve state of the art accuracy on English dependency parsing.
This approach corresponds to minimum Bayes risk decoding, and we conjecture that the arc attachment posterior marginals quantify a notion of uncertainty that may indicate difficulty or ambiguity.
Since running an ensemble is computationally expensive, we proposed discriminative training of a graph-based model with a novel cost function that \emph{distills} the ensemble uncertainty.   Deriving a cost function from a statistical model and extending distillation to structured prediction are new contributions.  This distilled model, trained to simulate the slower ensemble parser, improves over the state of the art on Chinese and German.

\section*{Acknowledgments}
We thank Swabha Swayamdipta, Sam Thomson, Jesse Dodge, Dallas Card, Yuichiro Sawai, Graham Neubig, and the anonymous reviewers for useful feedback. We also thank Juntao Yu and Bernd Bohnet for re-running the parser of \newcite{bohnet_nivre_12} on Chinese with gold tags. This work was sponsored in part by the Defense Advanced Research Projects Agency (DARPA) Information Innovation Office (I2O) under the Low Resource Languages for Emergent Incidents (LORELEI) program issued by DARPA/I2O under Contract No. HR0011-15-C-0114; it was also supported in part by Contract No. W911NF-15-1-0543 with the DARPA and the Army Research Office (ARO). Approved for public release, distribution unlimited. The views expressed are those of the authors and do not reflect the official policy or position of the Department of Defense or the U.S. Government. Miguel Ballesteros was supported by the European Commission under the contract numbers FP7-ICT-610411 (project MULTISENSOR) and H2020-RIA-645012 (project KRISTINA).


\bibliography{emnlp2016}
\bibliographystyle{emnlp2016}
\end{document}